\documentclass[sigconf]{acmart} 

\renewcommand\footnotetextcopyrightpermission[1]{} 

\usepackage{times}
\usepackage{epsfig}
\usepackage{graphicx}
\usepackage{amsmath}

\usepackage{amssymb}
\usepackage{enumerate}
\usepackage{algorithm}
\usepackage{algorithmic}
\usepackage{mathrsfs}
\usepackage[english]{babel}
\usepackage{array}
\usepackage{multirow}
\usepackage{color}
\usepackage{float}
\usepackage{booktabs}
\usepackage{enumitem}
\usepackage{balance}

\newcommand{\PreserveBackslash}[1]{\let\temp=\\#1\let\\=\temp}
\newcolumntype{C}[1]{>{\PreserveBackslash\centering}p{#1}}
\AtBeginDocument{%
  \providecommand\BibTeX{{%
    \normalfont B\kern-0.5em{\scshape i\kern-0.25em b}\kern-0.8em\TeX}}}

\acmSubmissionID{123-A56-BU3}

\copyrightyear{2021}
\acmYear{2021}
\setcopyright{acmcopyright}
\acmConference[MM '21] {Proceedings of the 29th ACM International Conference on Multimedia}{October 20--24, 2021}{Virtual Event, China.}
\acmBooktitle{Proceedings of the 29th ACM Int'l Conference on Multimedia (MM '21), Oct. 20--24, 2021, Virtual Event, China}
\acmPrice{15.00}
\acmISBN{978-1-4503-8651-7/21/10}
\acmDOI{10.1145/3474085.3475441}

\begin{document}
\fancyhead{}

\title{Unsupervised Image Deraining: Optimization Model Driven Deep CNN}

\author{Changfeng Yu}
\authornote{Both authors contributed equally to this research.}
\affiliation{%
  \institution{Huazhong University of Science and Technology}
  \city{Wuhan}
  \country{China}
  \postcode{430074}
}
\email{ycf@hust.edu.com}

\author{Yi Chang}
\authornotemark[1]
\affiliation{%
  \institution{Peng Cheng Laboratory}
  \city{Shenzhen}
  \country{China}
}
\email{owuchangyuo@gmail.com}

\author[a]{Yi Li}
\affiliation{%
  \institution{Huazhong University of Science and Technology}
  \city{Wuhan}
  \country{China}}
\email{li_yi@hust.edu.cn }

\author{Xile Zhao}
\affiliation{%
  \institution{University of Electronic Science and Technology of China}
  \city{Chengdu}
  \country{China}}
\email{xlzhao122003@163.com}

\author{Luxin Yan}
\authornote{Corresponding Author.}
\affiliation{%
  \institution{Huazhong University of Science and Technology}
  \city{Wuhan}
  \country{China}}
\email{yanluxin@hust.edu.cn}

\begin{abstract}
    The deep convolutional neural network has achieved significant progress for single image rain streak removal. However, most of the data-driven learning methods are full-supervised or semi-supervised, unexpectedly suffering from significant performance drop when dealing with the real rain. These data-driven learning methods are representative yet generalize poor for real rain. The opposite holds true for the model-driven unsupervised optimization methods. To overcome these problems, we propose a unified unsupervised learning framework which inherits the generalization and representation merits for real rain removal. Specifically, we first discover a simple yet important domain knowledge that \emph{directional rain streak is anisotropic while the natural clean image is isotropic}, and formulate the structural discrepancy into the energy function of the optimization model. Consequently, we design an optimization model driven deep CNN in which the unsupervised loss function of the optimization model is enforced on the proposed network for better generalization. In addition, the architecture of the network mimics the main role of the optimization models with better feature representation. On one hand, we take advantage of the deep network to improve the representation. On the other hand, we utilize the unsupervised loss of the optimization model for better generalization. Overall, the unsupervised learning framework achieves good generalization and representation: unsupervised training (loss) with only a few real rainy images (input) and physical meaning network (architecture). Extensive experiments on synthetic and real-world rain datasets show the superiority of the proposed method.
\end{abstract}


\maketitle

\section{Introduction}
    The single image rain streak removal \cite{kang2012automatic, luo2015removing, li2016rain, fu2017removing, yang2017deep, zhang2018density, Li2018Nonlocal, li2018recurrent, ren2019progressive, wang2019spatial, wei2019semi, hu2019depth, li2019heavy, yang2019joint, Yasarla2019Uncertainty, wang2019erl, Zhu2019Single, Wang2019DTDN, li2019single, Wang2020Model, Du2020Conditional, Yang2020Single, Deng2020Detail, Jiang2020Multi, Yasarla2020Syn} has made significant progress in the past decade, which serves as a pre-processing step for subsequent high-level computer vision tasks such as detection \cite{yolo9000} and segmentation \cite{zhao2017pyramid}. Most of the existing learning base CNN methods are full-supervised \cite{fu2017removing, yang2017deep} or semi-supervised \cite{wei2019semi, Yasarla2020Syn, Ye2021Closing}, which achieve satisfactory performance for the simulated rain streaks. However, the huge gap between the synthetic and real streaks would inevitably result in obvious performance drop. In this work, the goal is to handle the \emph{real rain streaks} from an unsupervised perspective.

\begin{figure*}
	\vspace*{-4.5mm}
\begin{center}
    \includegraphics[width=0.94\textwidth]{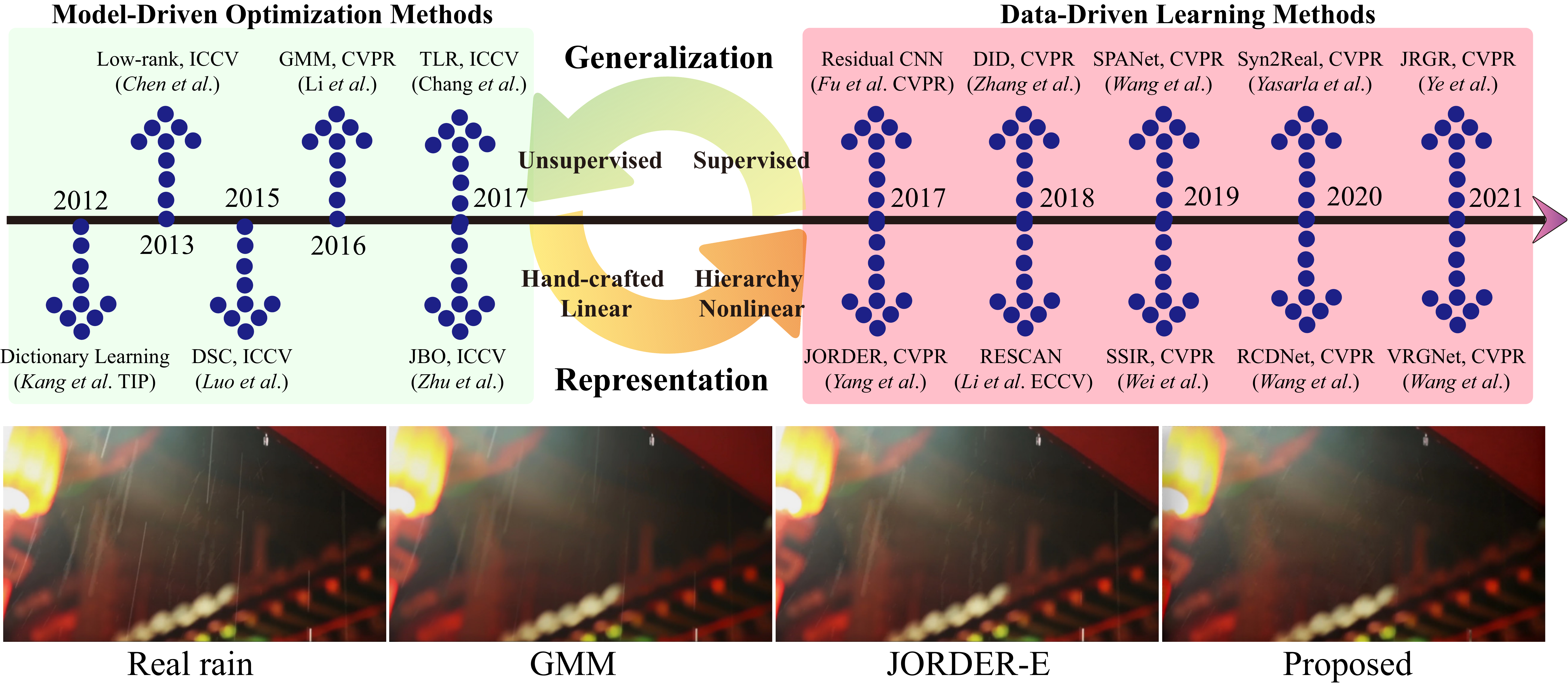}
 \end{center}
\vspace*{-3.5mm}
  \caption{The complementary between the optimization and CNN methods. We show the development of typical rain streak removal methods. The unsupervised model driven-optimization methods generalize well yet with only shallow representation. On the contrary, the full/semi-supervised based CNN deraining methods are representative with poor generalization. In this work, we bridge the gap between the model-driven and data-driven methods within an end-to-end unsupervised learning framework. Below shows the real rain removal results for representative optimization-based GMM \cite{li2016rain}, learning-based JORDER-E \cite{yang2019joint}, and the proposed method.}
  \label{Comparison}
\vspace*{-3.5mm}
\end{figure*}

    In pre-deep learning era, the optimization methods have achieved considerable progress in rain streaks removal. The main idea of optimization model is to formulate deraining task into an image decomposition framework by decoupling the rain streaks and clean image components which lie on two distinguishable subspaces. Thus, the key of optimization model is to construct an energy function and manually design hand-crafted priors for each component. The dictionary learning \cite{kang2012automatic, luo2015removing}, low-rank representation \cite{Chen2013A, chang2017transformed}, Gaussian mixture models (GMMs) \cite{li2016rain} have been widely explored for rain streaks removal. The optimization-based methods dig deeply into domain knowledge of the rain streaks, such as spatial sparsity \cite{kang2012automatic, Gu2017Joint}, smoothness \cite{zhu2017joint}, non-local similarity \cite{Chen2013A}, directionality \cite{chang2017transformed}. Moreover, they are usually free from the large-scale training datasets, so they can generalize well for real rain streaks. However, these hand-crafted priors are typically based on linear transformation, in which the representation ability is limited, especially for highly complex and varied rainy scenes. In addition, the optimization procedure is usually slow due to the multiple iterations procedure.

    Although the optimization-based methods have achieved promising deraining results, these hand-crafted priors are less robust to handle the rain streaks with diverse distributions, due to the varied angle, location, intensity, density, length, width and so on. In recent years, the deep learning based deraining methods \cite{fu2017removing, yang2017deep, zhang2018density, Li2018Nonlocal, li2018recurrent, ren2019progressive, wang2019spatial, wei2019semi, hu2019depth, li2019heavy, yang2019joint, Yasarla2019Uncertainty, wang2019erl, Zhu2019Single, Wang2019DTDN, li2019single, Wang2020Model, Du2020Conditional, Yang2020Single, Deng2020Detail, Jiang2020Multi, Yasarla2020Syn, Ye2021Closing} have received tremendous success in deraining task due to nonlinear representation ability of CNN. The powerful representation enables the CNN to implicitly learn different complex distribution of the rain streaks. Another advantage of the CNN methods is the fast inference time once the network is trained. The key components in conventional CNN are to 1) prepare the training data pairs; 2) design architecture of the network; 3) define loss function for the training purpose.

    Most of the existing CNN-based deraining methods are full-supervised, in which they pay most of their attention to the architecture design of the network such as the multi-stage \cite{yang2017deep, li2018recurrent, ren2019progressive, yang2019joint, Wang2020Model}, multi-scale \cite{pang2020single, Yasarla2019Uncertainty, Zhu2019Single, Jiang2020Multi}, attention \cite{Li2018Nonlocal, wang2019spatial, hu2019depth, Zhu2019Single}, so as to better improve the representation for the rain streaks. The full-supervised deraining methods heavily rely on the paired clean and rainy image. The existing full-supervised methods usually construct a rain synthesis model to generate the simulated rainy image. However, there exist a huge gap between the real and synthetic rains. That is the main reason why the existing CNN methods have been less generalized for the real rain streaks. The semi-supervised deraining methods \cite{wei2019semi, Yasarla2020Syn, Ye2021Closing} could alleviate the generalization issue to some extent by introducing the real rainy image as the additional constraint. However, the problem still exists since these semi-supervised methods also employ the synthetic rainy image.

    Overall, the model-driven optimization methods have good generalization endowed by the unsupervised loss yet weak representation (plane and linear) ability. On the contrary, the data-driven learning methods have good representation endowed by the hierarchy nonlinear transformation yet poor generalization (supervised loss on synthetic data) ability. This motivates us to inherit the powerful representation of the network and also the good generalization of optimization methods simultaneously.

    In this work, we bridge the gap between the model-driven and data-driven methods within an end-to-end unsupervised learning framework. Specifically, we first discover a simple yet important domain knowledge that \emph{directional rain streak is anisotropic while the natural clean image is isotropic}. This motivates us to construct a simple yet effective unsupervised directional gradient based optimization model (UDG) in which the rainy image is decomposed as the clean image regularized by isotropic TV and the rain streak constrained by anisotropic TV. UDG has good generalization yet poor representation ability and can be efficiently solved by the alternating direction method of multipliers \cite{lin2011linearized}. To further improve the representation of UDG, we design an UDG optimization model driven deep CNN (UDGNet). The architecture of the network mimics the main role of the optimization models with better feature representation. Consequently, the unsupervised loss of UDG is correspondingly enforced on UDGNet. Overall, the proposed method inherits good generalization and representation from both the optimization and CNN. The main contributions are summarized:

\begin{itemize}[leftmargin=1mm]
 \item Different from existing full/semi-supervised deraining methods, we attempt to solve real rain streaks from an unsupervised perspective. We connect the model-driven and data-driven methods via an unsupervised learning framework with simultaneous generalization and representation, which offers a new insight to deraining community.
  \item We discover the structural discrepancy between the rain streak and clean image. Consequently, we construct an unsupervised directional gradient based optimization model (UDG) for real rain streaks removal. Furthermore, we propose an optimization model-driven deep CNN (UDGNet) in which we optimize the network weights by minimizing the unsupervised loss function UDG.
  \item The proposed UDGNet can be trained with a few real rainy images, even one single image. We conduct extensive experiments on both synthetic and real-world datasets, which consistently perform superior against the state-of-the-art methods.
\end{itemize}

\begin{figure*}[t]
\begin{center}
    \includegraphics[width=0.905\textwidth]{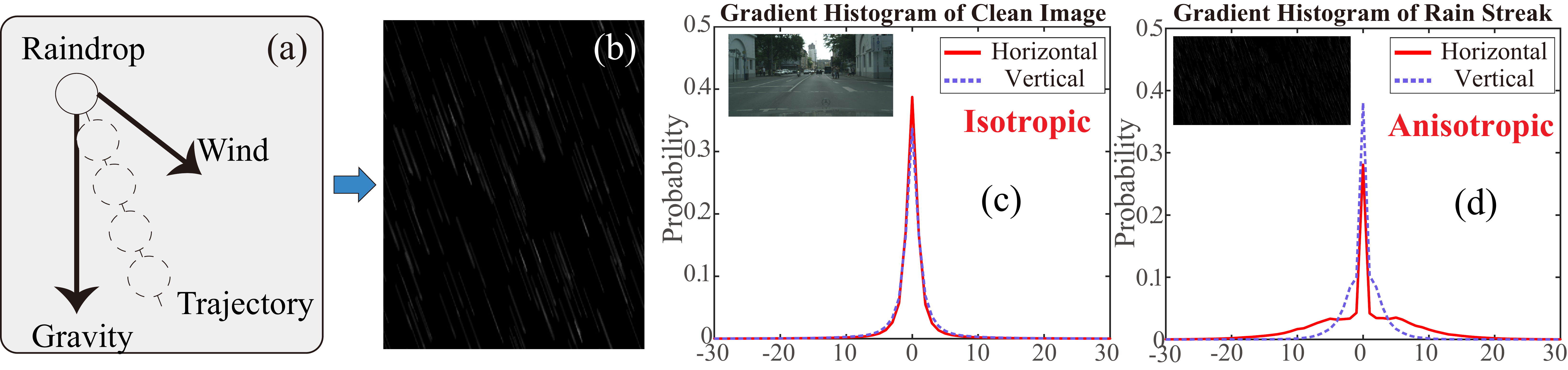}
\end{center}
  \vspace*{-3.5mm}
   \caption{The illustration of the rain streaks and the statistical discrepancy between the rain streak and clean image. (a) The analysis of the physical procedure of the raindrop in real-world space. (b) The visualization of the rain streaks in imaging space. (c) and (d) show the gradient histogram of the clean image and rain streak, respectively.}
\label{Structure Discrepancy}
  \vspace*{-3mm}
\end{figure*}

\vspace*{-2mm}
\section{Related Work}
\subsection{Optimization-based Deraining}
    The model-based optimization methods formulate the single image rain streaks removal task as an ill-posed problem, in which the decomposition framework is employed to model the image and rain streaks simultaneously with hand-crafted priors \cite{kang2012automatic, Chen2013A, luo2015removing, li2016rain, zhu2017joint, chang2017transformed, Gu2017Joint}. The key of the optimization method is to dig the domain-knowledge of both the rain streaks and image. In 2012, Kang \emph{et al}. \cite{kang2012automatic} excavated the spatial sparsity of both the image and rain streaks, and first introduced the one-dimensional vector-based dictionary learning with morphological component analysis. Later, Luo \emph{et al}. \cite{luo2015removing} proposed a discriminative sparse coding method by additionally forcing the two learned dictionaries with mutual exclusivity. Further, the authors utilized the non-local similarity of the images, and employed the two-dimensional low-rank matrix recovery \cite{Chen2013A} to better preserve the structure of the images. The directional property of the rain streak has been widely utilized. For example, Chang \emph{et al}. \cite{chang2017transformed} proposed a transformed low-rank model for compact rain feature representation. Li \emph{et al}. \cite{li2016rain} presented a simple patch-based Gaussian mixture models and can accommodate multiple orientations and scales of the rain streaks. In this work, we discover the structural discrepancy between the rain streak and clean image in the gradient domain, and propose a novel unsupervised directional gradient based optimization model for rain streaks removal. Moreover, we extend the proposed optimization model to the deep network by minimization of the unsupervised loss UDG, so as to significantly improve the feature representation for better real rain streaks removal.
\vspace*{-1mm}
\subsection{Learning-based Deraining}
    The CNN based single image rain streak removal methods can be mainly classified into the following categories: full-supervised \cite{fu2017removing, yang2017deep, zhang2018density, Li2018Nonlocal, li2018recurrent, ren2019progressive, wang2019spatial, hu2019depth, li2019heavy, yang2019joint, Yasarla2019Uncertainty, wang2019erl, Zhu2019Single, Wang2019DTDN, Wang2020Model, Du2020Conditional, Yang2020Single, Deng2020Detail, Jiang2020Multi}, semi-supervised \cite{wei2019semi, Yasarla2020Syn}, and unsupervised \cite{Zhu2019Single}. Most existing methods are full-supervised where the clean image and synthetic rain image pair are required. Fu \emph{et al}. \cite{fu2017removing} first introduced the end-to-end residual CNN to solve the rain streaks removal problem. Yang et al. \cite{yang2017deep} jointly detected and removed the rain in a multi-task network. The multi-stage and multi-scale architecture networks \cite{li2018recurrent, Wang2020Model, pang2020single, Yasarla2019Uncertainty, Jiang2020Multi} have been extensively studied for better feature representation. Ren \emph{et al}. \cite{ren2019progressive} proposed a simple yet effective progressive recurrent network with recursive blocks for image deraining. To better generalize the real rain streaks, the researchers employed the semi-supervised learning paradigm. For example, apart from the supervised loss, Wei \emph{et al}. \cite{wei2019semi} additionally enforced a parameterized GMM distribution on real rain streaks. To get rid of the limitation of the paired synthetic-clean training data, the unsupervised methods \cite{Zhu2019Single} have raised attention. The existing unsupervised methods all take advantage of the CycleGAN framework \cite{Zhu2017Unpaired} to handle the unpaired real rain streaks. In this work, our method starts from the unpaired and unsupervised network. To the best of our knowledge, we are the first unsupervised network that handles the real rainy image from the loss function perspective by utilizing the domain knowledge of the rain streaks.

\subsection{The Combination of Optimization and CNN}
    The model-driven optimization methods and the data-driven learning methods are the two main categories restoration methods over the past decades. These two methodologies are complementary to each other in terms of the generalization, representation, training and testing time. There are many attempts to combine them into a unified framework. The most popular way is the plug-and-play strategy \cite{zhang2017learning}. Benefiting from the variable splitting techniques \cite{lin2011linearized}, the discriminative CNN can be plugged into model-based restoration methods as a learnable regularization. Liu \emph{et al}. \cite{Liu2018Deep} exploited a deep layer prior under the \emph{maximum-a-posterior} framework to recover the intrinsic rain structure. Another typical manner is the unfolding \cite{yang2016deep}, which designs the network with sufficient interpretability by unfolding the iterative optimization procedure into a deep network architecture. Wang \emph{et al}. \cite{Wang2020Model} designed a rain convolutional dictionary RCDNet for image deraining with exact step-by-step corresponding relationship between the network modules and the operators in optimization procedure. In this work, we unify the optimization model and CNN into an end-to-end unsupervised learning network, in which we optimize the network weights by minimizing an unsupervised loss function, derived from the anisotropic smoothness knowledge of rain streaks. Moreover, the architecture of the model-driven deep neural network is interpretable with better feature representation.

\begin{figure*}[t]
	\vspace*{-3mm}
\begin{center}
    \includegraphics[width=0.92\textwidth]{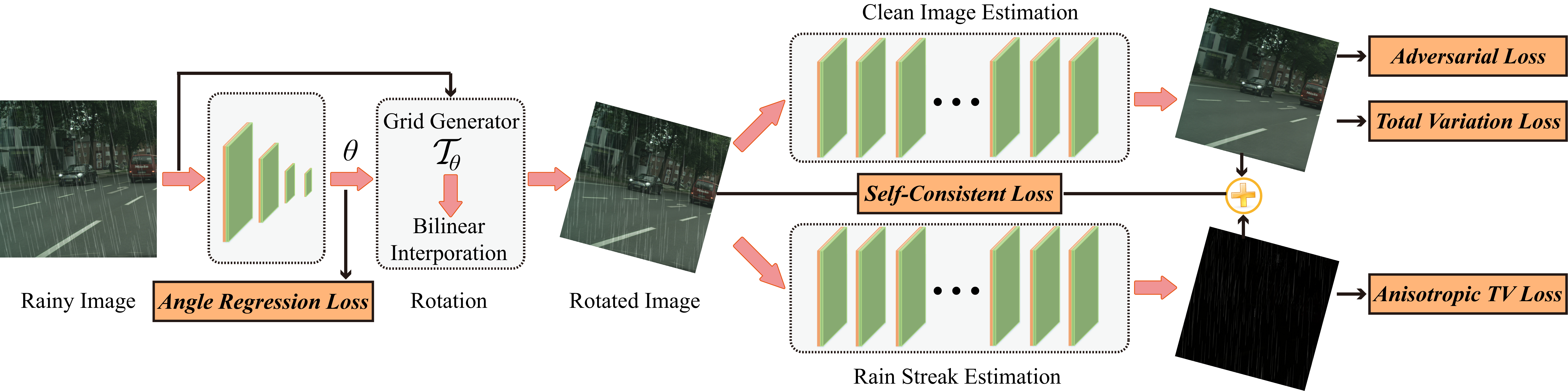}
\end{center}
\vspace*{-3mm}
   \caption{The architecture of the proposed network. The UDGNet mimics the main operations in optimization procedure. Given the rainy image, we first estimate the principal direction angle of the rain streaks and feed it to the simplified spatial transformation module \cite{jaderberg2015spatial} so as to obtain the regular vertical rain streaks. Then, the rotated image is decomposed into distinguishable image and rain streak subspaces, which is realized by the domain knowledge driven unsupervised isotropic and directional anisotropic gradient loss. Finally, we reconstruct the rotated rainy image with the self-consistency loss to further improve the feature representation.}
\label{Network}
\vspace*{-3mm}
\end{figure*}

\section{Methodology}
\subsection{Discrepancy Between Image and Rain Layer}
    The key of the unsupervised optimization method is to excavate the domain knowledge of both the clean image and rain streaks, so as to decouple the two components into distinguishable subspaces. The sparsity, low-rank, GMM properties have been extensively utilized in previous works. In this work, we discover a simple yet important domain knowledge that directional rain streak is anisotropic while the natural clean image is isotropic\footnote{Note that in this work, we relax the isotropic to the horizontal and vertical direction.}.
Here, we provide an analysis to support our statement from a physical and image statistical viewpoint. On one hand, the physical shape of the rain is approximately spherical raindrop \cite{garg2007vision}. The raindrops are affected by both the gravity and wind. The gravity ensures that the rain is vertically descending and the wind determines the descending angle, as shown in Fig. \ref{Structure Discrepancy}(a). Then, the imaging system maps the raindrop of the real world to the image plane. Due to the long exposures of the imaging and fast motion of the raindrops, the visual appearance of the rain in image space is presented as severely motion-blurred rain streaks, as shown in Fig. \ref{Structure Discrepancy}(b). That is to say, the directional rain streaks are naturally anisotropic. On the contrary, the natural image is typically vertical and horizontal isotropic \cite{torralba2003statistics}.

    We further statistically demonstrate that the structure discrepancy between clean image and line-pattern rain streaks. Specifically, we first calculate the gradient maps (first-order forward difference) along both vertical and horizontal axis, and then calculate the histogram of gradient maps (\emph{x-axis} denotes the gradient bins, \emph{y-axis} represents the number count) on large-scale datasets, as shown in Fig. \ref{Structure Discrepancy}(c) and (d). The isotropic means that properties are the same when measured along axes in different directions. The gradient histograms of the image along different directions including \emph{x} and \emph{y} axis are very close to each other (isotropic TV), while this property does not hold true for the rain streaks (anisotropic TV). That is to say, the directional property of the rain streaks mainly increases the gradient variation across the streak line direction while has less influence along the streak line. The structure discrepancy between the clean image and the rain streak is the key to decouple them into two subspaces.

\subsection{Unsupervised Directional Gradient Model}
    Now, the key problem is how to mathematically formulate the structure discrepancy into the optimization model. As for clean image, we utilize isotropic total variational to depict the isotropic gradient smoothness along both horizontal and vertical dimension. As for rain streak, we would like to design an anisotropic directional gradient constraint: penalize the gradient along the rain streak while preserving the gradient across rain streak. This is very reasonable, since the rain streaks exhibit obvious directionality similar to the sharp edges. However, the main difficulty is the arbitrary angle of rain streaks in different images. To solve this problem, we follow the rotated model in \cite{chang2017transformed}, so as to obtain vertical rain streaks:
\begin{equation}\label{eq:DegradationModel}
\setlength{\abovedisplayskip}{0.1mm}
\setlength{\belowdisplayskip}{0.1mm}
\theta\circ\textbf{\emph{Y}} = \textbf{\emph{X}} + \textbf{\emph{R}},
\end{equation}
    where \textbf{\emph{Y}} is the observed rainy image, $\theta$ is the rotation angle, \textbf{\emph{X}} is the clean background, and \textbf{\emph{R}} is the rain streaks. The goal of this work is to estimate both clean image \textbf{\emph{X}} and rain streaks \textbf{\emph{R}} simultaneously from the given rainy image \textbf{\emph{Y}}. The general optimization deraining model can be deduced via the \emph{maximum-a-posterior} as follow:
\begin{equation}\label{eq:RegularizationFormulation}
\setlength{\abovedisplayskip}{0.5pt}
\setlength{\belowdisplayskip}{0.5pt}
\mathop {min}\limits_{\textbf{\emph{X}},\textbf{\emph{R}}} \frac{1}{2}||\textbf{\emph{X}} + \textbf{\emph{R}} - \theta\circ\textbf{\emph{Y}}||_F^2 + \tau P_{x}(\textbf{\emph{X}}) + \lambda P_{r}(\textbf{\emph{R}}),
\end{equation}
where the first term is the data fidelity, $P_{x}$ and $P_{r}$ denote the prior term on clean image and rain streaks, respectively. According to the analysis above, we choose the conventional isotropic TV for the clean image and anisotropic directional TV for the rain streaks. For simplicity of the optimization, we estimate the angle of the $\theta$ via TILT \cite{zhang2012tilt} in advance, and denote $\textbf{\emph{Y}}_r = \theta\circ\textbf{\emph{Y}}$. Thus, the formulation of the unsupervised directional gradient image deraining model is:
\begin{equation}\label{eq:Original LRTV Model}
\setlength{\abovedisplayskip}{0.5pt}
\setlength{\belowdisplayskip}{0.5pt}
\mathop {min}\limits_{\textbf{\emph{X}},\textbf{\emph{R}}} \frac{1}{2}||\textbf{\emph{X}} + \textbf{\emph{R}} - \textbf{\emph{Y}}_r||_F^2 + \tau ||\textbf{\emph{X}}|{|_{\emph{TV}}} + \lambda ||\textbf{\emph{R}}|{|_{\emph{UTV}}},
\end{equation}
where $||\textbf{\emph{X}}|{|_{\emph{TV}}} =  ||\nabla \textbf{\emph{X}}||_1$ and $||\textbf{\emph{R}}|{|_{\emph{UTV}}} =  {||{\nabla _x} \textbf{\emph{R}}||_{1} + ||{{\nabla _y} \textbf{\emph{Y}}_r - {\nabla _y}\textbf{\emph{R}}}||_1}$,  and $\nabla  = ({\nabla _x};{\nabla _y})$ denotes the vertical (along the rain streak) and horizontal  (across the rain streak) derivative operators, respectively, where $||\cdot|{|_1}$ denotes the sum of absolute value of the matrix elements.

1) \textbf{Rain Streaks Update}: given image \textbf{\emph{X}}, the rain streaks \textbf{\emph{R}} can be estimated from the following minimization problem:
\begin{equation}\label{eq:Stripe Estimation Problem}
\setlength{\abovedisplayskip}{1pt}
\setlength{\belowdisplayskip}{1pt}
\resizebox{0.919\hsize}{!}{$\hat{\textbf{\emph{R}}} = \arg \mathop {\min }\limits_\textbf{\emph{R}} \frac{1}{2}||\textbf{\emph{X}} + \textbf{\emph{R}} - \textbf{\emph{Y}}_r||_F^2 + \lambda_x ||{\nabla _x} \textbf{\emph{R}}||_{1} + \lambda_y ||{{\nabla _y} \textbf{\emph{R}} - {\nabla _y}\textbf{\emph{Y}}_r}||_1.$}
\end{equation}
Due to the non-differentiability of the ${L_1}$ norm, we introduce the ADMM \cite{lin2011linearized} to convert the original problem into two easy subproblems with closed-form solutions. By introducing two auxiliary variables ${\textbf{\emph{P}}_x} = {\nabla _x}\textbf{\emph{R}}$ and ${\textbf{\emph{P}}_y} = {{\nabla _y} \textbf{\emph{R}} - {\nabla _y}\textbf{\emph{Y}}_r}$, the Eq. (\ref{eq:Stripe Estimation Problem}) is equivalent to following problem:
\begin{equation}\label{eq:Equivalent Problem}
\setlength{\abovedisplayskip}{0mm}
\setlength{\belowdisplayskip}{0mm}
\begin{small}
\begin{aligned}
&\{\hat{\textbf{\emph{R}}},\hat{\textbf{\emph{P}}}_x,\hat{\textbf{\emph{P}}}_y\}  = \mathop {\arg\min }\limits_{\textbf{\emph{R}},{\textbf{\emph{P}}_x},{\textbf{\emph{P}}_y}} \frac{1}{2}||\textbf{\emph{X}} + \textbf{\emph{R}} - \textbf{\emph{Y}}_r||_F^2 +  \lambda_x ||{\textbf{\emph{P}}_x}|{|_1} +  \lambda_y ||{\textbf{\emph{P}}_y}|{|_1}\\
& +\frac{\alpha }{2}||{\textbf{\emph{P}}_x} - {\nabla _x}\textbf{\emph{R}} - \frac{{{\textbf{\emph{J}}_x}}}{\alpha }||_F^2 + \frac{\beta }{2}||{\textbf{\emph{P}}_y} - ({\nabla _y}\textbf{\emph{R}} - {\nabla _y}\textbf{\emph{Y}}_r) - \frac{{{\textbf{\emph{J}}_y}}}{\beta }||_F^2.
\end{aligned}
\end{small}
\end{equation}

$\blacksquare$ The \textbf{\emph{R}}-related subproblem is
\begin{equation} \label{eq:R-Subproblem}
\begin{small}
\begin{aligned}
\setlength{\abovedisplayskip}{0mm}
\setlength{\belowdisplayskip}{0mm}
\hat{\textbf{\emph{R}}} = \arg \mathop {\min }\limits_{\textbf{\emph{R}}} &\frac{1}{2}||\textbf{\emph{X}} + \textbf{\emph{R}} - \textbf{\emph{Y}}_r||_F^2  + \frac{\alpha }{2}||{\textbf{\emph{P}}_x} - {\nabla _x}\textbf{\emph{R}} - \frac{{{\textbf{\emph{J}}_x}}}{\alpha }||_F^2\\
&+ \frac{\beta}{2}||{\textbf{\emph{P}}_y} - ({\nabla _y}\textbf{\emph{R}} - {\nabla _y}\textbf{\emph{Y}}_r) - \frac{{{\textbf{\emph{J}}_y}}}{\beta }||_F^2,
\end{aligned}
\end{small}
\end{equation}
which has a close-form solution via 2-D fast Fourier transform (FFT)
\begin{equation} \label{eq:R-Solution}
\resizebox{0.89\hsize}{!}{${\textbf{\emph{R}}^{k + 1}} = {\mathcal{F}^{ - 1}}\left( {\frac{{\mathcal{F}\left( {(\textbf{\emph{Y}}_r - \textbf{\emph{X}}^{k}) + {\nabla_x ^T}(\alpha^{k} {\textbf{\emph{P}}_x^{k}} - {\textbf{\emph{J}}_x^{k}}) + {\nabla_y ^T}(\beta^{k} {\textbf{\emph{P}}_y^{k}} + \beta^{k}{\nabla _y}\textbf{\emph{Y}}_r - {\textbf{\emph{J}}_y^{k}}) } \right)}}{{1 + \alpha^{k} {{\left( {\mathcal{F}(\nabla_x)} \right)}^2} + \beta^{k} {{\left( {\mathcal{F}(\nabla_y)} \right)}^2}}}} \right).$}
\end{equation}
\vspace*{-1mm}
$\blacksquare$ The $\{\textbf{\emph{P}}_x, \textbf{\emph{P}}_y\}$-related subproblem is
\begin{equation} \label{eq:P-Subproblem}
\setlength{\abovedisplayskip}{0.5mm}
\setlength{\belowdisplayskip}{0.5mm}
\left\{ {\begin{array}{*{20}{c}}
 {\hat{\textbf{\emph{P}}_x}  = \arg \mathop {\min }\limits_{\textbf{\emph{P}}_x} \lambda_x ||\textbf{\emph{P}}_x|{|_1} + \frac{\alpha }{2}||\textbf{\emph{P}}_x - \nabla_x \textbf{\emph{R}} - \frac{\textbf{\emph{J}}_x}{\alpha }||_F^2}\\
 \resizebox{0.84\hsize}{!}{${\hat{\textbf{\emph{P}}_y}  = \arg \mathop {\min }\limits_{\textbf{\emph{P}}_y} \lambda_y ||\textbf{\emph{P}}_y|{|_1} + \frac{\beta }{2}||\textbf{\emph{P}}_y - ({\nabla _y}\textbf{\emph{R}} - {\nabla _y}\textbf{\emph{Y}}_r) - \frac{\textbf{\emph{J}}_y}{\beta }||_F^2}.$}
 \end{array}} \right.
\end{equation}
which can be solved efficiently via a soft shrinkage operator:
\begin{equation} \label{eq:P-Solution}
\setlength{\abovedisplayskip}{0.5pt}
\setlength{\belowdisplayskip}{0.5pt}
\left\{ {\begin{array}{*{30}{c}}
{{\textbf{\emph{P}}_x^{k + 1}} = shrink\_{\emph{L}_1}(\nabla_x {\textbf{\emph{R}}^{k + 1}} + \frac{{{\textbf{\emph{J}}_x^k}}}{\alpha^{k} },\frac{\lambda_x }{\alpha^{k} })}\\
{{\textbf{\emph{P}}_y^{k + 1}} = shrink\_{\emph{L}_1}(\nabla_y {\textbf{\emph{R}}^{k + 1}} - {\nabla _y}\textbf{\emph{Y}}_r + \frac{{{\textbf{\emph{J}}_y^k}}}{\beta^{k} },\frac{\lambda_y }{\beta^{k} })}\\
{shrink\_{\emph{L}_1}(r,\xi ) = \frac{r}{{|r|}}*\max (|r| - \xi ,0).}
\end{array}} \right.
\end{equation}

Finally, the Lagrangian multipliers and penalization parameters are updated as follows:
\begin{equation} \label{eq:Multipliers-Updata}
\left\{ {\begin{array}{*{30}{c}}
{\textbf{\emph{J}}_x^{k + 1}} = {\textbf{\emph{J}}_x^k} + {\alpha ^k}(\nabla_x {\textbf{\emph{R}}^{k + 1}} - {\textbf{\emph{P}}_x^{k + 1}})\\
{\textbf{\emph{J}}_y^{k + 1}} = {\textbf{\emph{J}}_y^k} + {\beta ^k}(\nabla_y {\textbf{\emph{R}}^{k + 1}} - \nabla_y {\textbf{\emph{Y}}_r} - {\textbf{\emph{P}}_y^{k + 1}})\\
\{{\alpha ^{k + 1}}, {\beta ^{k + 1}}\}= \{{\alpha ^k}, {\beta ^k}\}\cdot \rho.
\end{array}} \right.
\end{equation}

2) \textbf{Image Update}: given rain streak \textbf{\emph{R}}, the image \emph{\textbf{X}} can be estimated from the following minimization problem:
\begin{equation} \label{eq:Image Estimation Problem}
\setlength{\abovedisplayskip}{0.5pt}
\setlength{\belowdisplayskip}{0.5pt}
\hat{\textbf{\emph{X}}} = \arg \mathop {\min }\limits_\textbf{\emph{X}} \frac{1}{2}||\textbf{\emph{X}} + \textbf{\emph{R}} - \textbf{\emph{Y}}_r||_F^2 + \tau ||{\nabla }\textbf{\emph{X}}|{|_1}.
\end{equation}
The optimization of Eq. (\ref{eq:Image Estimation Problem}) is similar to that of Eq. (\ref{eq:Stripe Estimation Problem}). Here, we do not describe the procedure in detail.

\begin{figure}[t]
	
\begin{center}
    \includegraphics[width=0.48\textwidth]{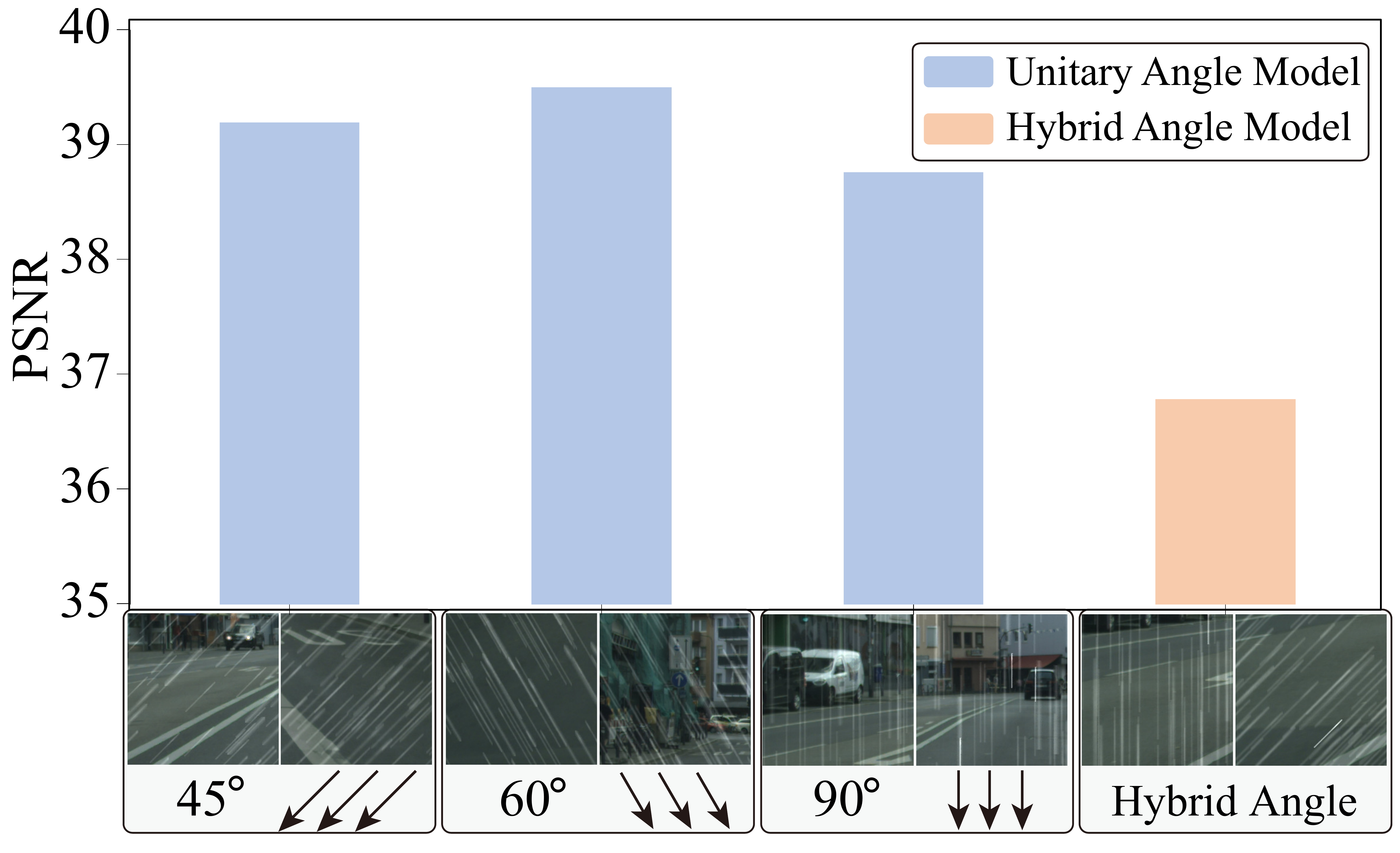}
\end{center}
\vspace*{-4mm}
   \caption{The advantage of rotation for rainy images. The same model trained on the unitary rain streak angle dataset significantly outperforms that of the hybrid angle.}
   \vspace*{-5mm}
\label{Adavantage of Rotation}
\end{figure}

\subsection{The Optimization Model-driven Deep CNN}
Although the UDG could remove most of the rain streaks and well accommodate different rainy images, the residual rain streaks and over-smooth phenomenon are easily observed, especially for the complex scenes. The main reason is the limited representation ability of the hand-crafted linear transformation of TV prior. In this work, we bridge the gap between the optimization model and the deep CNN to overcome this issue.

From the optimization model perspective, we deepen the shallow optimization model by introducing deep CNN with powerful representation ability. Specifically, we introduce CNN to approximate clean image and rain streaks, and enforce the unsupervised loss of optimization model as constraint for the deep CNN. Thus, the unsupervised learning framework can retain generalization ability, meanwhile leverage the hierarchy nonlinear representation of CNN.

From the deep learning perspective, we replace the conventional supervised paired constraint by the unsupervised loss of the optimization model. Thus, we get rid of the paired clean-synthetic labels for supervised training and directly train from the real rainy images. The unsupervised loss endows us the good generalization ability with powerful representation of the network. Moreover, most of the optimization methods are time-consuming, since the iteration of the solving large-scale linear systems are required. Thus, the proposed method can enjoy a fast inference speed of CNN.

\noindent
\textbf{The Architecture and Loss of UDGNet}.
Now we introduce concrete architecture of the proposed network which is used for minimizing the defined unsupervised loss function. The overall network architecture  in Fig. \ref{Network} is to mimic the optimization procedure of Eq. (\ref{eq:RegularizationFormulation}). Specifically, we first learn the rotation angle $\theta$ regression module which repeatedly includes several conv and pooling layers.
\begin{equation}\label{Angel Loss}
\setlength{\abovedisplayskip}{1pt}
\setlength{\belowdisplayskip}{1pt}
{\mathcal{L}_{{\theta}}={\left\Vert {{\theta} -  \mathcal{F}\left( \textbf{\emph{Y}};\textbf{\emph{W}}_{\theta} \right)} \right\Vert}_{{2}}},
\end{equation}
where \textbf{\emph{Y}} is the input rainy image, $\mathcal{F}(\bullet)$ is the network transformation and $\textbf{\emph{W}}_{\theta}$ is the learnable angle regression parameters, and ${\theta}$ is ground-truth of the rain streaks which can be easily obtained in advance. Compared with the 2D image and rain streak, the angle is a single scalar, which is much easier to learn and label.

\begin{figure*}[t]
\begin{center}
    \includegraphics[width=0.94\textwidth]{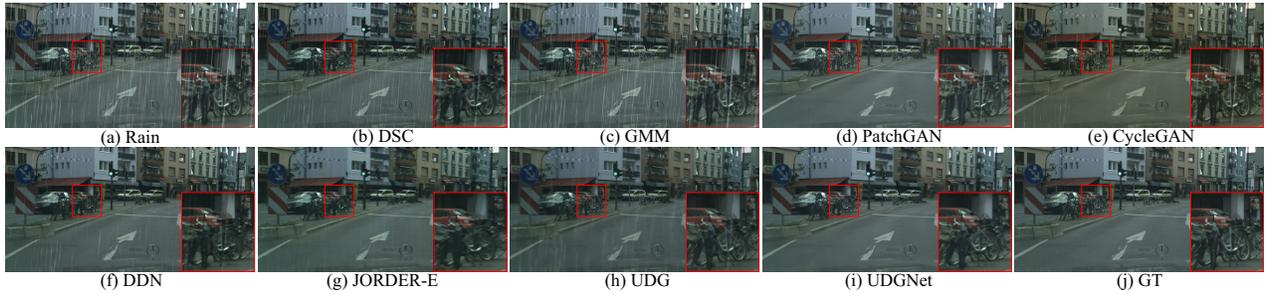}
\end{center}
\vspace*{-4mm}
   \caption{Visualization of deraining results on Cityscape dataset.}
\label{Cityscape}
\vspace*{-3mm}
\end{figure*}

\begin{figure*}[t]
\begin{center}
    \includegraphics[width=0.94\textwidth]{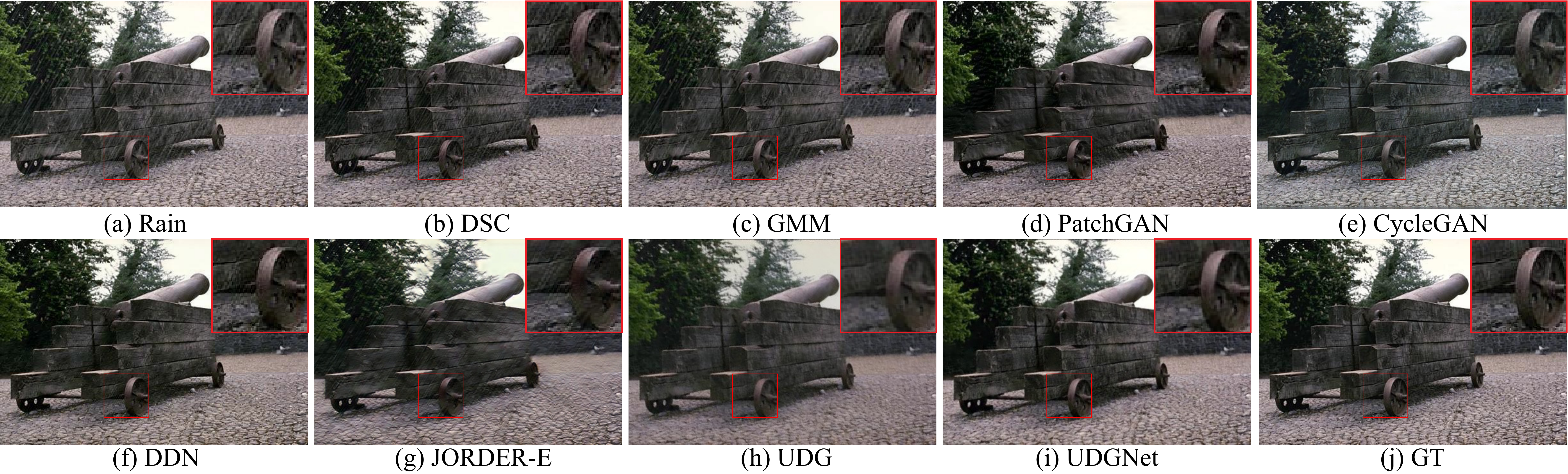}
\end{center}
\vspace*{-4mm}
   \caption{Visualization of deraining results on Rain1400 dataset.}
\label{Rain1400}
\vspace*{-3mm}
\end{figure*}

\begin{table*}[]
\renewcommand{\arraystretch}{1.05}
\centering
\small
\caption{Quantitative comparison PSNR and SSIM with state-of-the-art methods on synthetic datasets.}
\vspace*{-3mm}
\begin{tabular}{c|c|c|c|c|c|c|c|c|c|c|c|c}
\hline
    \multirow{2}{*}{Dataset}               &    \multirow{2}{*}{Index}    &    \multirow{2}{*}{Rain}     & \multicolumn{3}{c|}{Optimization} & \multicolumn{3}{c|}{Full-supervised} & Semi & \multicolumn{3}{c}{Unsupervised} \\ \cline{4-13}
&  &    & DSC       & GMM       & UDG       & DDN        & JORDER-E   & RCDNet    & SSIR            & GAN   & Cyclegan  & UDGNet  \\ \hline
\multirow{2}{*}{Cityscape} & PSNR  & 26.22  &     28.18      &    29.11       & 26.95     & 30.59      & 24.76      & 28.87     &     25.15            & 30.09      & 31.62      & \textbf{34.65}   \\ \cline{2-13}
                           & SSIM  & 0.7776 &   0.8104        &   0.8649        & 0.9434    & 0.9232     & 0.8559      & 0.8778    &     0.8165            & 0.9325     & 0.9140     & \textbf{0.9653}  \\ \hline
\multirow{2}{*}{Rain1400}  & PSNR  & 23.68  & 26.26     & 25.72     & 23.35     & 28.07      & 22.18       & 24.69     &    25.67             & 24.14      & 28.32      & \textbf{29.16}   \\ \cline{2-13}
                           & SSIM  & 0.7542 & 0.7828    & 0.7797    & 0.8380    & 0.8633     & 0.7719      & 0.7840    &     0.8183            & 0.7880     & 0.8685     & \textbf{0.8910}  \\ \hline
\end{tabular}
\label{Simulated Quantitative}
\vspace*{-3mm}
\end{table*}

Next, we feed the learned scalar angle into a modified differentiable spatial transform module \cite{jaderberg2015spatial}, which explicitly allows the spatial affine transformation of the input image. The classical STN \cite{jaderberg2015spatial} can not control how the transformed feature map is. Compared with the classical STN, we additionally enforce a physical meaning rotation angle ${\theta}$, so that the rainy image \textbf{\emph{Y}} with arbitrary direction can be well rectified into the vertical direction $\textbf{\emph{Y}}_r$. We show the original image \textbf{\emph{Y}} and the rotated version $\textbf{\emph{Y}}_r$, in Fig. \ref{Network}. Such a simple operation would significantly reduce the rain streak removal difficulty by reducing the angle variations of different rain streaks. To show the effectiveness of the rotation, we train on two cases: rain streaks with arbitrary directions and rain streaks with only the vertical direction. The results are reported in Fig. \ref{Adavantage of Rotation}, which is not surprising since the rotation has explicitly eliminated the uncertainty.

Then, we construct two parallel streams for the image and rain streaks estimation, analog to the two prior terms $P_{x}(\textbf{\emph{X}})$ and $P_{r}(\textbf{\emph{R}})$ in Eq. (\ref{eq:RegularizationFormulation}). Each stream corresponds to the alternating minimization of Eq. (\ref{eq:Stripe Estimation Problem}) and Eq. (\ref{eq:Image Estimation Problem}) in optimization model. Furthermore, we additionally introduce the adversarial loss \cite{Goodfellow2014Generative} on the clean image for better textures preserving. The architecture of the two streams are the same with 32 Resblocks \cite{he2016deep}. Thus, instead of directly minimization of the clean image and the rain streaks, we learn the parameters $\textbf{\emph{W}}_{I}$ and $\textbf{\emph{W}}_{R}$ in each stream as follow:
\begin{equation}\label{Image Loss}
\setlength{\abovedisplayskip}{0.5mm}
\setlength{\belowdisplayskip}{0.5mm}
{\mathcal{L}_{{image}} = \tau{ \left\Vert { \nabla \mathcal{F}\left( \textbf{\emph{Y}}_{{r}};\textbf{\emph{W}}_{{I}} \right) } \right\Vert}_{{1}}} + \mu\mathcal{L}_{adv},
\end{equation}
\begin{equation}\label{Rain Loss}
\setlength{\abovedisplayskip}{0.5mm}
\setlength{\belowdisplayskip}{0.5mm}
\resizebox{0.90\hsize}{!}{${\mathcal{L}_{{rain}}= \lambda_x{ \left\Vert { \nabla_x \mathcal{F}\left( \textbf{\emph{Y}}_{{r}};\textbf{\emph{W}}_{{R}} \right) } \right\Vert}_{{1}} + \lambda_y{ \left\Vert { \nabla_y \mathcal{F}\left( \textbf{\emph{Y}}_{{r}};\textbf{\emph{W}}_{{R}} \right) - \nabla_y\textbf{\emph{Y}}_{{r}}} \right\Vert}_{{1}} },$}
\end{equation}
where $\mathcal{F}(\bullet)$ is the network transformation, $\mathcal{L}_{adv}$ is the adversarial loss \cite{Goodfellow2014Generative}. The first term TV loss in Eq. (\ref{Image Loss}) serves as the local pixel-level smoothness prior while the second term adversarial loss works as the global image-level statistical prior. The two terms are complementary to each other, so as to obtain natural and clean image.

Finally, we enforce the self-consistency constraint by composing the estimated image and rain streaks back to the rotated rainy image, which is exactly the first data fidelity term in Eq. (\ref{eq:RegularizationFormulation}):
\begin{equation}\label{Self-consistency Loss}
\setlength{\abovedisplayskip}{1pt}
\setlength{\belowdisplayskip}{1pt}
{\mathcal{L}_{{self}}={\left\Vert { \textbf{\emph{Y}}_{{R}} -  \mathcal{F}\left( \textbf{\emph{Y}}_r;\textbf{\emph{W}}_{I} \right) -  \mathcal{F}\left( \textbf{\emph{Y}}_r;\textbf{\emph{W}}_{R} \right)} \right\Vert}_{{2}}}.
\end{equation}
Thus, the overall loss of the proposed network can be interpreted as:
\begin{equation}\label{Overall Loss}
\setlength{\abovedisplayskip}{1pt}
\setlength{\belowdisplayskip}{1pt}
\mathcal{L}_{{overall}}= \mathcal{L}_{{\theta}} + \mathcal{L}_{{image}} + \mathcal{L}_{{rain}} + \mathcal{L}_{{self}}.
\end{equation}

The deep network explicitly learns the optimization procedure, in which each module and loss mimic the main operation in optimization. On one hand, the network inherits the unsupervised domain knowledge from optimization. Thus, the proposed network generalizes well to different rainy images and can be trained and tested on one single image (without the adversarial loss). Moreover, once the network is trained, it only requires a very fast forward pass through the deep network to predict the clean image without further optimization steps. On the other hand, the proposed model is representative of the complex scenes endowed by the highly-nonlinear network, and the proposed network is interpretable and controllable.

\begin{figure*}
	\vspace*{-4mm}
\begin{center}
    \includegraphics[width=0.90\textwidth]{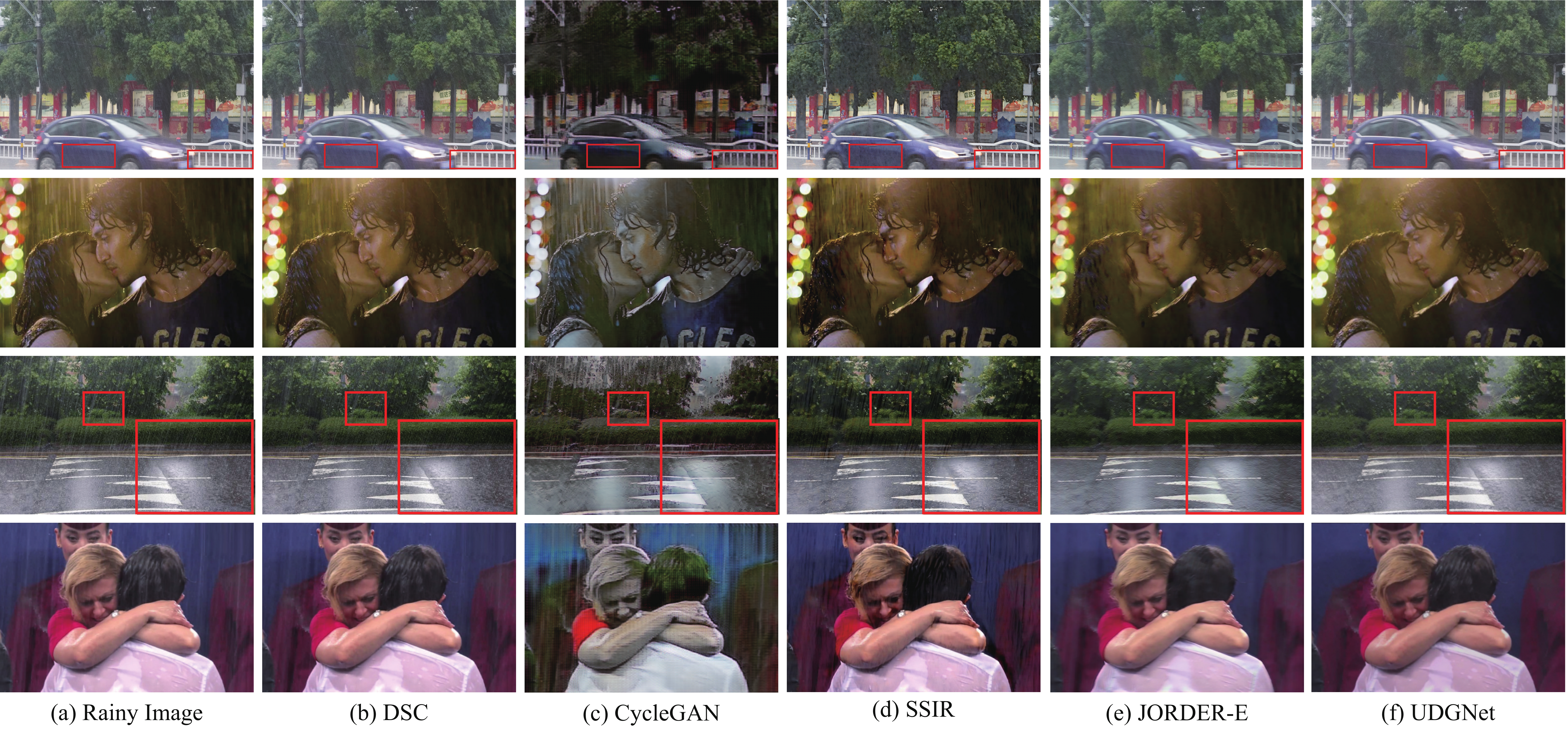}
\end{center}
\vspace*{-4mm}
   \caption{Visualization of deraining results on RealRain dataset.}
\label{RealRain}
\vspace*{-3mm}
\end{figure*}

\begin{table}[]
\renewcommand{\arraystretch}{1.05}
\centering
\small
\caption{Quantitative comparison NIQE and User study results with state-of-the-art methods on real dataset.}
\vspace*{-3.5mm}
\begin{tabular}{c|c|c|c|c}
\hline
Method & Rainy    & DSC    & GMM      & UDG    \\ \hline
NIQE   & 5.70     & 5.55   & 6.06     & 5.12    \\ \hline
User study & 1.00 & 2.88	& 3.64	&4.53		\\ \hline
Method & JORDER-E & RCDNet & CycleGAN & UDGNet \\ \hline
NIQE   & 5.48     & 5.62   & 6.54     & 5.08         \\ \hline
User study & 4.03 & 3.03	& 1.14	&5.69		\\ \hline
\end{tabular}
\label{Real Quantitative}
\vspace*{-3mm}
\end{table}

\section{EXPERIMENTAL RESULTS}
\subsection{Datasets and Experimental Setting}
\noindent \textbf{Datasets.} We evaluate UDGNet on both synthetic and real datasets.
\begin{itemize}[leftmargin=*]
\item \textbf{Cityscape.} We simulate cityscapes rain images following the screen blend model \cite{luo2015removing} with different streak length, width, angle and intensity. We follow the cityscapes dataset with 2975 images for training and 500 images for testing.
\item \textbf{Rain1400.} We adopt rain1400 \cite{fu2017removing} as another synthetic dataset, which contains 1400 rain/clear images pairs. We randomly select 12600  for training and 1400 pairs for testing.
\item \textbf{RealRain.} We collect real world rainy images with large field of view from the datasets \cite{yang2017deep, wei2019semi, chen2018robust} and Google search. We utilize 100 real rainy images to train and test the UDGNet.
\end{itemize}

\begin{table}[]
\renewcommand{\arraystretch}{1}
\centering
\small
\caption{Model size (MB) and time complexity (seconds)}
\vspace*{-3mm}
\begin{tabular}{c|c|c|c|c}
\hline
Method & DSC    & GMM    & UDG      & DDN    \\ \hline
Model size   & -     & -   & -     & 0.233    \\ \hline
Running time & 5066 & 1443	& 7.846	&0.162		\\ \hline
Method & JORDER-E & RCDNet & CycleGAN & UDGNet \\ \hline
Model size   & 16.7     & 13.1   & 11.7     & 5.7         \\ \hline
Running time & 9.123 & 22.6	& 3.098	&0.129		\\ \hline
\end{tabular}
\label{Model parameter and complexity}
\vspace*{-5mm}
\end{table}

\par\noindent \textbf{Implemention Details.} The images are trained and tested through sliding windows with the size of 128*128. The angle of rain streaks is obtained via the TILT \cite{zhang2012tilt} in advance for training. We adopt Adam \cite{Kingma2014Adam} as the optimizer with batch size of 8. The initial learning rate is set to be 0.001 and decay 0.1 every 30 epochs. We set the hyper-parameter $\tau$, $\mu$, $\lambda_x$, $\lambda_y$ as 0.01, 400, 1.5, 1.0, respectively.

\noindent \textbf{Experimental Setting.} We compare the proposed unsupervised UDG and UDGNet with (1) optimization methods DSC \cite{luo2015removing} and GMM \cite{li2016rain}; (2) supervised methods DDN \cite{fu2017removing}, JORDER-E \cite{yang2019joint} and RCDNet \cite{Wang2020Model}; (3) semi-supervised methods SSIR \cite{wei2019semi}; (4) unsupervised network PatchGAN \cite{Isola2017Image} and CycleGAN \cite{Zhu2017Unpaired}. For synthetic data, the full-reference PSNR and SSIM are utilized as the quantitative evaluation. For real-world images, we employ the non-reference NIQE \cite{Mittal2013Making} and user studies to quantitatively evaluate the visual quality of deraining results. The higher PSNR, SSIM and user study point is and the lower the NIQE is, the better the deraining result is. The optimization methods do not need the training dataset. The supervised methods are trained on the defined dataset in the original paper and tested on different datasets in our work.

\begin{figure*}
		\vspace*{-4mm}
	\begin{center}
		\includegraphics[width=0.94\textwidth]{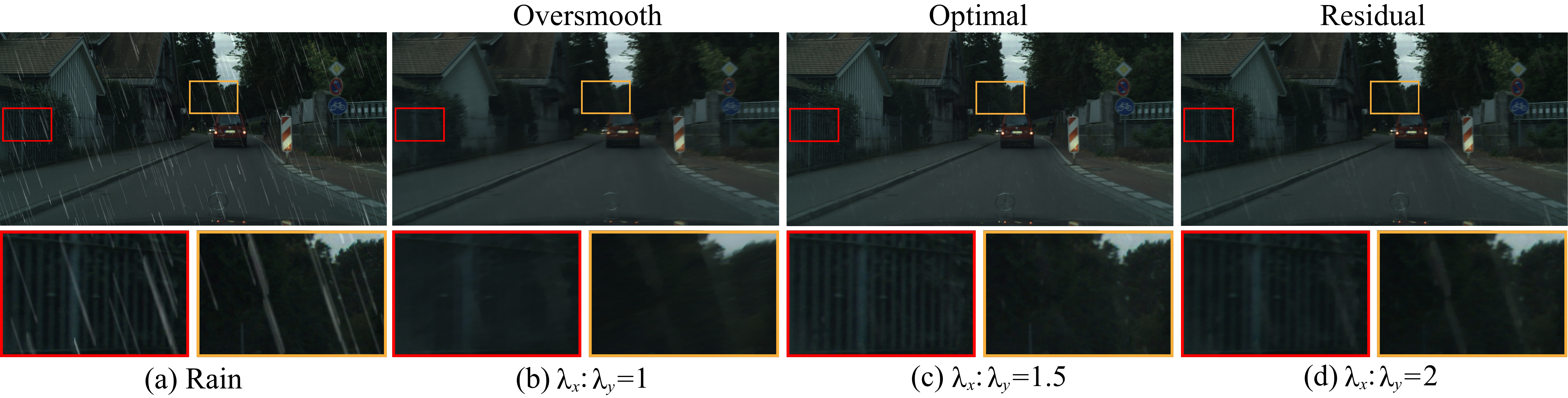}
	\end{center}
	\vspace*{-3mm}
	\caption{The illustration of how the regularization parameters control the deraining strength.}
	\label{Regularization Parameters}
	\vspace*{-3.5mm}
\end{figure*}

\begin{figure}
	\begin{center}
		\includegraphics[width=0.45\textwidth]{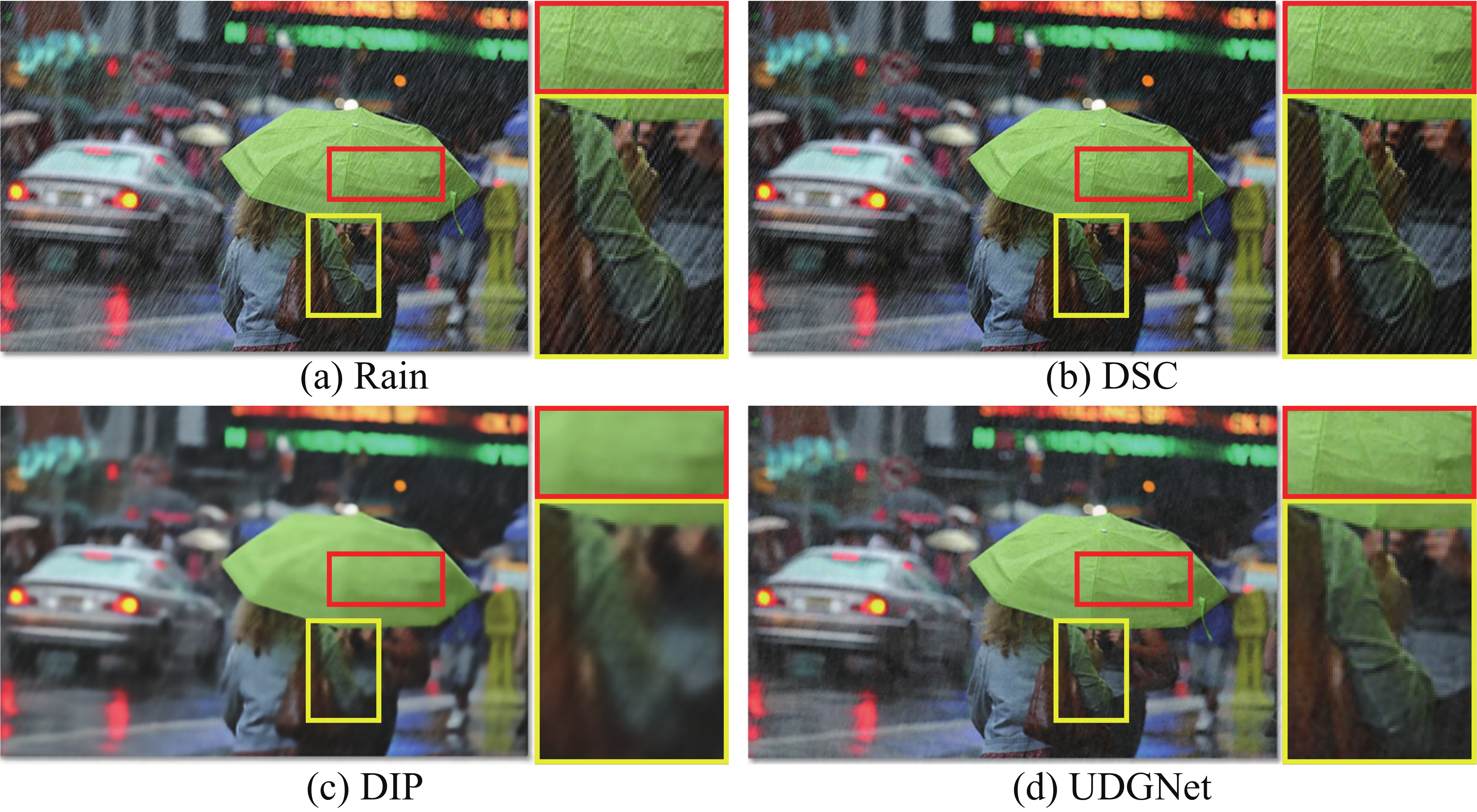}
	\end{center}
	\vspace*{-3.5mm}
	\caption{The single image training and inferencing.}
	\label{Single Image Training}
	\vspace*{-3mm}
\end{figure}

\begin{table}[]
	\renewcommand{\arraystretch}{1.05}
	\centering
	\small
	\caption{The influence of the estimated angle $\theta$.}
	\vspace*{-3.5mm}
	\begin{tabular}{c|c|c|c|c|c}
		\hline
		$\theta$ & Estimated & GT     & GT$\pm$5$^\circ$   & GT$\pm$10$^\circ$  & GT$\pm$20$^\circ$  \\ \hline
		PSNR  & 34.65     & 34.71  & 34.56  & 34.26  & 33.03  \\ \hline
		SSIM  & 0.9663    & 0.9665 & 0.9639 & 0.9602 & 0.9389 \\ \hline
	\end{tabular}
	\label{angle estimation}
	\vspace*{-3.6mm}
\end{table}

\subsection{Quantitative and Qualitative Results}

\noindent\textbf{Qualitative Results.}
In Fig. \ref{Cityscape}-\ref{RealRain}, we show the visual deraining results on both synthetic and real datasets: Cityscape, Rain1400 and RealRain. We can observe that there are obvious residual rain streaks in optimization-based methods, especially in Fig. \ref{Cityscape}, because of the limited representation ability of hand-crafted priors for diverse background and rain streaks. The unsupervised learning-based methods PatchGAN and CycleGAN could remove most of the rain streaks with a few residual. However, GAN-based unsupervised methods are difficult to train and easy to collapse, since they heavily rely on the distribution of training dataset. The supervised methods DDN and JORDER-E could suppress rain streaks to some extent, while there are also some residual rain streaks in the results due to the data distribution discrepancy. The optimization model UDG could remove most rain streaks. UDGNet further enhances the performance. On one hand, the visual rain streaks have been satisfactorily removed by the UDGNet with few residuals. On the other hand, compared with the UDG, the unsupervised learning-based UDGNet could well preserve the image structures. Compared with other methods, the proposed UDGNet could achieve better performance in terms of both rain streaks removal and image texture preserving.

\begin{table}[]
\renewcommand{\arraystretch}{1.05}
\centering
\small
\caption{The effectiveness analysis of each loss in UDGNet.}
\vspace*{-3.5mm}
\begin{tabular}{C{0.55cm}|C{1.1cm}|C{1.3cm}|C{0.75cm}|C{0.75cm}|C{0.85cm}|C{0.85cm}}
\hline
Case &$\mathcal{L}_{img-tv}$ & $\mathcal{L}_{img-adv}$ &  $\mathcal{L}_{rain}$ & $\mathcal{L}_{self}$ & PSNR & SSIM \\ \hline
1&$\surd$     &  $\times$     &  $\times$     & $\times$      &    22.63   &   0.8229   \\ \hline
2&$\times$    &  $\surd$    &   $\times$    &  $\times$      &  30.09    &  0.9325    \\ \hline
3&$\surd$     &  $\surd$     & $\times$      &  $\times$      &   30.08   &   0.8915   \\ \hline
4&$\times$    &  $\times$    &  $\surd$     &  $\times$      &    32.44  &  0.9538    \\ \hline
5&$\times$   &   $\surd$     & $\surd$       &   $\surd$     &  33.01    &  0.9681    \\ \hline
6&$\surd$    &  $\times$     &   $\surd$    &  $\surd$     &  34.64    &  0.9644    \\ \hline
7&$\surd$    &  $\surd$      &  $\surd$      &  $\surd$     &   34.65     &  0.9653    \\ \hline
\end{tabular}
\label{Ablation Loss}
\vspace*{-3.8mm}
\end{table}

\noindent\textbf{Quantitative Results.}
The quantitative results are reported in Table \ref{Simulated Quantitative} and \ref{Real Quantitative} in which the best results are in bold. The UDGNet consistently obtains the best deraining results, which further demonstrates the superior of the proposed method in terms of the performance and generalization. We further show model size and time complexity of different methods in Table \ref{Model parameter and complexity}.
The proposed UDGNet is computationally cheap and efficient. The model size of UDGNet is about 5.7MB, which is significantly smaller than the competing methods. Furthermore, we benchmark the running time with an Intel Core i7-8700 CPU and an NVIDIA RTX 2080Ti. For an image with size 1024*2048, the running time of the UDG is 7.8s, which is obviously faster than DSC (5066s) and GMM (1433s). The testing time of UDGNet is 0.12s, which is more attractive for practical use.
\vspace*{-1mm}
\subsection{Ablation Study}
\noindent
\textbf{How does Angle Estimation Affect the Performance?}
The angle estimation and rotation module is an important pre-processing part of our UDGNet. In Table \ref{angle estimation}, we show how the angle estimation influences the deraining performance. We can observe that the more accurate the angle is, the better the deraining result is, which indicates that precise angle guidance can indeed improve UDGNet for better deraining performance. Moreover, the deraining result of the estimated angle is very close to that of the provided the oracle (GT) angle, which validates the effectiveness of the proposed network.


\noindent
\textbf{What is the Effectiveness of Each Loss?}
To verify the effect of each loss in UDGNet, in Table \ref{Ablation Loss}, we conduct ablation study of each term on cityscape validation. From cases 1, 2, 3: the adversarial loss is more important than the TV loss for image. From case 4: the proposed directional domain knowledge of rain streak is very effective for rain removal. From case 5, 6, 7: the joint loss with both rain streak and clean image could further boost the performance. Moreover, the self-consistency loss is also beneficial to the performance.


\noindent
\textbf{How Can We Control the Deraining Result of UDGNet?}
The loss function of UDGNet is derived from optimization model (UDG), in which each term has clear interpretability. Thus the deraining strength can be controlled through hyper-parameters $\lambda_x$ and $\lambda_y$. As shown in Fig. \ref{Regularization Parameters}, different deraining results are obtained with different $\lambda_x/\lambda_y$ ratios. As ratio increases, more details are preserved with more streaks left, and vice versa. Thus, we can set different ratios to balance texture preservation and rain streak removal adaptively.

\begin{figure}

	\begin{center}
		\includegraphics[width=0.47\textwidth]{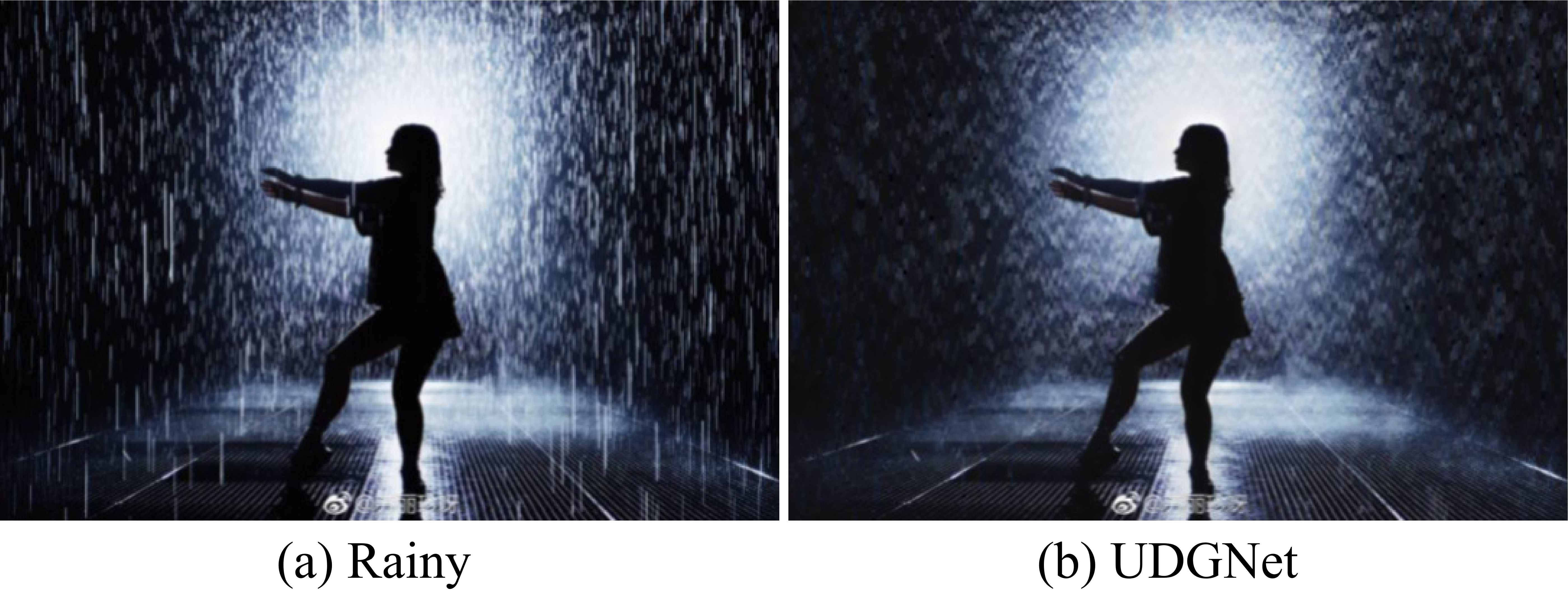}
	\end{center}
	\vspace*{-3.5mm}
	\caption{The limitation of the UDGNet.}
	\label{Limitation}
	\vspace*{-3.5mm}
\end{figure}

\subsection{Discussion}
\noindent
\textbf{Single Image Training and Inference}
The previous learning based methods consistently need a number of the training samples, in which the testing performance heavily relies on the training datasets. In this work, our unsupervised learning framework not only utilizes external dataset but also the internal prior knowledge of singe image. Therefore, the UDGNet can be trained on both the large scale datasets and one single image. We test the performance of single image training along with the similar deep image prior DIP \cite{ulyanov2018deep} for comparison in Fig. \ref{Single Image Training}. We can observe that the UDGNet could well remove and rain streaks with clear image texture, while the DIP has unexpectedly over-smooth the image details.

\noindent
\textbf{Limitation}
In Fig. \ref{Limitation}, we show a real image with rain streaks (nearby) and veiling artifacts (distant). We can observe that the rain streaks have been satisfactorily removed while the veiling effects still exist in the result. This is reasonable, since the UDGNet mainly utilizes the directional anisotropic characteristic in the loss. In future, we would like to incorporate the domain knowledge of the veiling artifacts via the unsupervised loss into the proposed framework.

\section{Conclusion}
In this work, we aim at the real image rain streak removal, and propose a novel optimization model driven deep CNN method for unsupervised deraining. Our start point is to bridge the gap between the model-driven optimization method and the data-driven learning method in terms of the generalization and representation. The key to our learning framework is the modelling of the structure discrepancy between the rain streak and clean image. We formulate this domain knowledge into unsupervised direction gradient optimization model, and transfer the unsupervised loss function to the deep network, such that the proposed method could simultaneously achieve good generalization and representation ability. Extensive experimental results on both the real and synthetic datasets demonstrate the effectiveness of the proposed method for real rain streaks removal.

\textbf{Acknowledgements.}This work was supported by National Natural Science Foundation of China under Grant No. 61971460, China Postdoctoral Science Foundation under Grant 2020M672748, National Postdoctoral Program for Innovative Talents BX20200173 and  Equipment Pre-Research Foundation under grant No. 6142113200304.
\balance
\bibliographystyle{ACM-Reference-Format}
\bibliography{acmart}

\end{document}